# A Fuzzy-based Framework to Support Multicriteria Design of Mechatronic Systems

Abolfazl Mohebbi,[a,1] Sofiane Achiche,[a] Luc Baron [a]

[a] *Department of Mechanical Engineering, Polytechnique Montreal, Montreal, Quebec, Canada.*

**Abstract**

Designing a mechatronic system is a complex task since it deals with a high number of system components with multi-disciplinary nature in the presence of interacting design objectives. Currently, the sequential design is widely used by designers in industries that deal with different domains and their corresponding design objectives separately leading to a functional but not necessarily an optimal result. Consequently, the need for a systematic and multi-objective design methodology arises. A new conceptual design approach based on a multi-criteria profile for mechatronic systems has been previously presented by the authors which uses a series of nonlinear fuzzy-based aggregation functions to facilitate decision-making for design evaluation in the presence of interacting criteria. Choquet fuzzy integrals are one of the most expressive and reliable preference models used in decision theory for multicriteria decision making. They perform a weighted aggregation by the means of fuzzy measures assigning a weight to any coalition of criteria. This enables the designers to model importance and also interactions among criteria thus covering an important range of possible decision outcomes. However, specification of the fuzzy measures involves many parameters and is very difficult when only relying on the designer's intuition. In this paper, we discuss three different methods of fuzzy measure identification tailored for a mechatronic design process and exemplified by a case study of designing a vision-guided quadrotor drone. The results obtained from each method are discussed in the end.

---

[1] Corresponding author at abolfazl.mohebbi@polymtl.ca

## List of symbols

| | |
|---|---|
| $MMP$ | Mechatronic Multicriteria Profile |
| $GCS$ | Global Concept Score |
| $MIQ$ | Machine Intelligence Quotient |
| $RS$ | Reliability Score |
| $CX$ | Design Complexity |
| $FX$ | Design Flexibility |
| $CT$ | Cost of manufacture and production |
| $m_i$ | Criteria values |
| $\phi_i$ | Normalized sub-criteria values |
| $\mu$ | Fuzzy measures |
| $\lambda$ | Sugeno measure |
| $I$ | Interaction index |
| $\phi$ | Importance index |
| $C_\mu$ | Choquet Integral |
| $E$ | Error criterion |
| **u** | A vector containing all the coefficients of fuzzy measures |

# 1. Introduction

Multidisciplinary systems that include synergetic integration of mechanical, electrical, electronic, and software components, are known as Mechatronic Systems (Rzevski 2014). Because of the high number of the constituent components, the multi-physical aspect of the subsystems and the couplings between the different engineering disciplines involved, the design of mechatronic systems can be rather complex and it requires an integrated and concurrent approach to obtain optimal solutions (Torry-Smith, Qamar et al. 2013, Mohebbi, Baron et al. 2014). In a similar manner to other systems, the design of mechatronic devices includes three major phases: conceptual design, detailed design, and prototyping and improvements. Several problems and limitations are encountered when the design is at its early stages, as it requires choosing the "Elite Set" which is the selection of components and choosing between alternatives for software and control strategies. This practice creates challenges due to insufficient support of the multi-criteria nature of mechatronics systems design, which calls for decision support across various disciplines. In such cases, design engineers tend to choose the first and the best components from what they see as available and feasible to meet their design requirements. Such decisions can often lead to a functional design, but rarely to an optimal one. This ill decision making generally occurs due to improperly-defined performance criteria and lack of knowledge about the co-influences between criteria and the functionality to be provided by neighboring disciplines.

The present paper contributes towards a better concept evaluation process during the conceptual design phase. The goal of concept evaluation is to compare the generated concepts based on the design requirements and to select the best alternative for further device and then product development. (Tomiyama, Gu et al. 2009) presented a comprehensive description of the design theory and methodology (DTM) and an evaluation of its application in practical scenarios. (Ullman 1992) has analyzed four concept evaluation methods. All of these methods provide qualitative frameworks to evaluate candidate solutions. The results of these comparisons highly depend on the experience of the design engineer. Novice designers would make decisions easier if quantitative evaluation methods are available for them. To this effect, an evaluation index can be used to rank the generated feasible solutions and therefore more easily choose between design alternatives. (Moulianitis, Aspragathos et al. 2004) introduced a mechatronic index that characterizes the mechatronic designs by their control performance, complexity, and flexibility. The overall evaluation was formulated based on the averaging operators and weight factors were manually applied to highlight the importance of each criterion. They did not, however, consider the interactions between design criteria. (Behbahani and de Silva 2007) proposed a framework for the design of mechatronic systems in which the performance requirements were represented by a mechatronic design quotient (MDQ). Correlations between design criteria have been taken into account by using fuzzy functions. MDQ was implemented in a number of case studies (Behbahani and de Silva 2008), and was claimed to be efficient; however, the assessment of criteria was very qualitative and no systematic measurement approach has been presented nor implemented, which puts the burden on the engineering designers.

(Mohebbi, Achiche et al. 2014) presented a new approach based on their newly introduced multi-criteria mechatronic profile (MMP) for the conceptual design stage. The MMP included five main elements of machine intelligence, reliability, flexibility, complexity, and cost, while each main criterion has several sub-criteria. To facilitate fitting the intuitive requirements for decision-making in the presence of interacting criteria, three different criteria aggregation methods were proposed and inspected using a case study of designing a vision-guided quadrotor drone and also a robotic visual servoing system. These methods benefit from three different aggregation techniques namely: Choquet integral, Sugeno integral (Mohebbi, Achiche et al. 2014), and a fuzzy-based neural network (Mohebbi, Baron et al. 2014). These techniques proved to be more precise and reliable in multi-criteria design problems where interaction between the objectives cannot, and should not, be overlooked (Moghtadernejad, Chouinard et al. 2018, Moghtadernejad, Chouinard et al. 2020). The Choquet integral is one of the most expressive preference models used in decision theory. It performs a weighted aggregation of criteria using a capacity function assigning a weight to any coalition of criteria. This enables the expression of both positive and negative interactions and covering an important range of possible decision dilemmas, which is generally ignored in other multicriteria decision making (MCDM) methods (Grabisch 1996, Grabisch 1997). A 2-additive Choquet integral has been used in (Mohebbi, Achiche et al. 2014), which only uses relatively simple quadratic complexity and enables the modeling of the interaction between pairs of criteria.

Despite the modeling capabilities, the specification of the fuzzy measures has been always a place for various challenges which makes the practical use of such aggregation techniques difficult. While the definition of a simple weighted sum operator with $n$ criteria requires $n-1$ parameters, the definition of the Choquet integral with $n$ criteria requires setting of $2^n - 2$ capacities (measures), which can become quickly unmanageable even for low values of $n$ and even for an expert who can assess the coefficients based on semantical considerations. Most of the previous works on the capacity specification for Choquet integral-based decision analysis, consider a static preference database as input (learning set) and focuses on the determination of a set of measures that best fits the available preferences (Marichal and Roubens 2000). For example, a quadratic error between Choquet values and target utility values prescribed by the decision-maker (DM) can be minimized on a sample of reference alternatives (Meyer and Roubens 2006). Generally, questions are asked to the decision-maker, and the information obtained is represented as linear constraints over the set of parameters. An optimization problem is then solved to find a set of parameters that minimizes the error according to the information given by the decision-maker (Grabisch 1995). In (Marichal and Roubens 1998), it is supposed that an expert is able to tell the relative importance of criteria and identify the type of interaction between them if any. These relations can be expressed as a partial ranking of the alternatives on a global basis; partial ranking of the criteria, partial ranking of interaction indices and also the type of interaction between some pairs of criteria. These approaches differ with respect to the optimization objective function and the preferential information they require as input. (Rowley, Geschke et al. 2015) and (Moghtadernejad, Mirza et al. 2019) proposed methods to extract the fuzzy measures using the Principal Component Analysis (PCA). The method is based on identifying a measure of independence among design criteria. Two major problems of the aforementioned approaches are the lack of transparency on how the measures are made, the lack of robustness and the lack of reproducibility (Timonin 2013). Another alternative seems to be appropriate when using an optimization algorithm alongside a minimal intuitive determination by the decision-maker. These approaches take advantage of the lattice structure of the coefficients (Mori and Murofushi 1989).

While most of these methods are developed within a pure mathematical framework, some others were reflected in a limited number of applications such as computer vision, pattern recognition, software engineering, and website design. To our knowledge, none of the developed approaches are applied to an inherently cross-disciplinary engineering design problem with multiple design objectives, e.g. mechatronics design. In this paper, we will explore various approaches of fuzzy measure identification applied to a mechatronic design problem. A Choquet integral aggregation was previously used by the authors for the multicriteria design of a mechatronic system in (Mohebbi, Achiche et al. 2014, Mohebbi, Achiche et al. 2019) where the measures were determined intuitively by the authors and a group of 30 researchers (all specialized in system design and mechatronics) through a questionnaire. The presented paper is organized as follows: Section 2 gives a brief overview of the conceptual design of mechatronic systems and the previously developed methodology based on the Mechatronic Multicriteria Profile (MMP) as a design evaluation index. Fuzzy decision support and the Choquet aggregation technique are described in Section 3 alongside the necessary definitions on fuzzy measures and integrals, illustrated with some properties. Section 4 describes three different algorithms for elicitation and identification of fuzzy measures with their philosophy, while Section 5 reports the results of a case study to incorporate and compare all the design evaluation attempts. Finally, Section 6 discusses the concluding remarks of the presented research.

## 2. Multicriteria Design of Mechatronic Systems

### 2.1. Conceptual Design

Conceptual design is an early stage of design in which the designers generally choose amongst the concepts that fulfill the design requirements and then decide how to interconnect these concepts into system architectures. Usually, at the beginning of every conceptual design process, a large number of candidate concepts exist for a given design problem. Consequently, a considerable amount of uncertainty arises about which of these solutions will be best fitted to the given requirements and objectives. This is more evident when the designer has to meet highly dynamic and interconnected design requirements. It is crucial to abandon the traditional end-to-end and sequential design process and to consider all aspects of a design problem concurrently. This is particularly necessary for multi-disciplinary systems such as mechatronic systems where mechanical, control, electronic, and software components interact and a high-quality design cannot be achieved without simultaneously considering all domains (Rzevski 2003).

## 2.2. Concept Evaluation

To achieve more optimal mechatronics designs, one requires a systematic evaluation approach to choose amongst the candidate design solutions. This evaluation includes both comparison and decision making (Coelingh, de Vries et al. 2002). In other words, decision-making is achieved by selecting the "best" alternatives by comparison. It is crucial to take into account both correlation between system requirements and also interactions between the multidisciplinary subsystems. The candidate solutions are generated based on a series of design specifications, candidate solutions are generated. The goal of concept evaluation is to compare the generated concepts against the requirements and to select the best one for the detailed design and optimization stages. This process is illustrated in Figure 1.

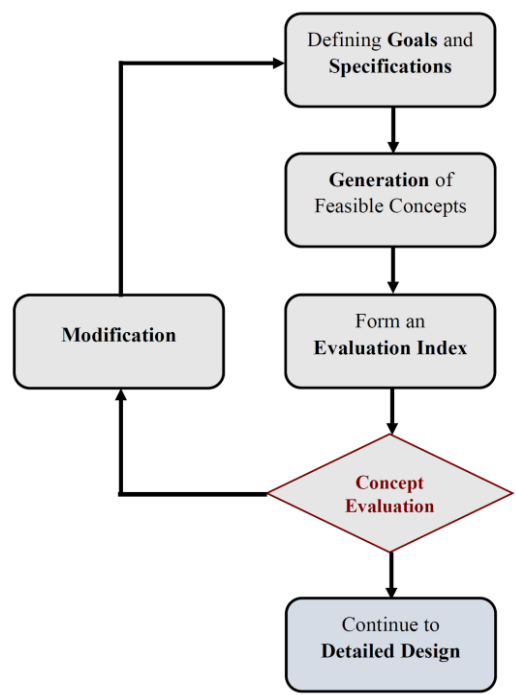

Fig. 1: Process of concept evaluation in design

## 2.4. Mechatronic Multicriteria Profile (MMP)

One important challenge faced during conceptual design is to find the right set of criteria to concurrently evaluate and synthesize the designs. Generally, making design decisions with multiple criteria is often performed using a Pareto approach. Without the identification of the system performance parameters and the full understanding of their co-influences, it is unrealistic to expect achieving optimal solutions. In order to form an integrated and systematic evaluation approach, the most important criteria and their related sub-criteria have been quantified by the authors in (Mohebbi, Achiche et al. 2014) to form an index vector of five normalized elements called Mechatronic Multicriteria Profile (MMP) as follows:

$$MMP = [MIQ, RS, CX, FX, CT]^T \qquad (1)$$

where *MIQ* is the machine intelligence quotient, *RS* is the reliability score, *CX* is the design complexity, *FX* is the flexibility, and *CT* is the cost of manufacture and production. Figure 2 describes the MMP with all corresponding sub-criteria. MMP will be used in this paper. We also define $x_i$ as the parameters used in calculating a criterion $i$, using which the criteria values are calculated using a function $f$ and $0 \leq f(x_i) \leq 1$. After determination and normalizing each sub-criterion, and by using a linear summation of weighted factors, the value of each main criterion will be assessed as follows:

$$f(x_i) = \sum_{j=1}^{n} w_j \bar{\rho}_i \qquad (2)$$

where $\bar{\rho}_i$ is the calculated value for each sub-criterion, $n$ is the total number of sub-criteria, and $w_j$ are the assigned-by-designer weights associated with each sub-criterion.

### *2.5. Detailed Design*

Preliminary features of a structure and the architecture of the mechatronic system are decided in the conceptual design stage where the components and subsystems of the product are specified. The control scheme is also selected in this stage without specifying its parameters. Subsequently, the calculation and specification of design parameters are done in the detailed design stage. Some of the design parameters can be specified or tuned after the machine is built (Realtime parameters or RTPs) and some others are not (Non-RTP). Regardless of these categories, all design variables should be computed and optimized in a concurrent and integrated manner concerning multiple criteria that affect the performance of the system. We previously proposed an integrated approach for the detailed design of mechatronic systems formulated in a multi-objective cross-disciplinary design optimization problem in which the design objectives of all subsystems are considered alongside the corresponding constraints (Mohebbi, Achiche et al. 2019). This approach is summarized in Figure 3.

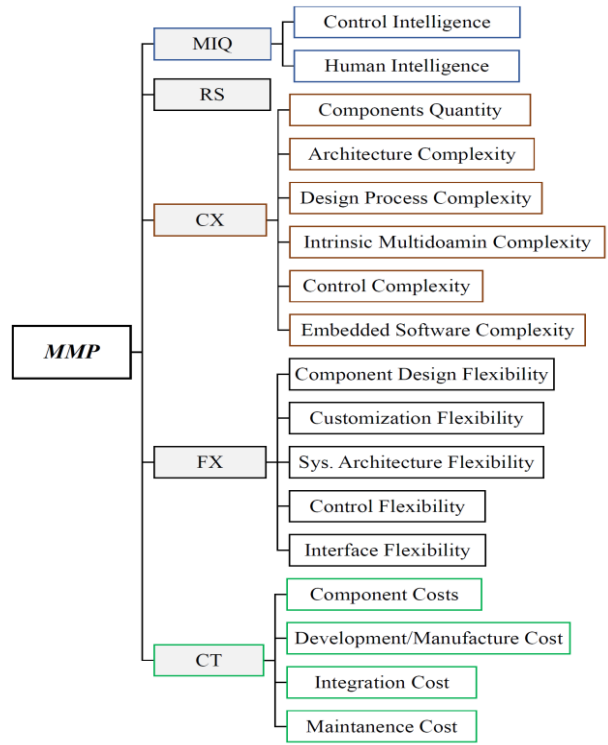

Fig. 2: Mechatronic Multicriteria Profile (MMP) and all sub-criteria

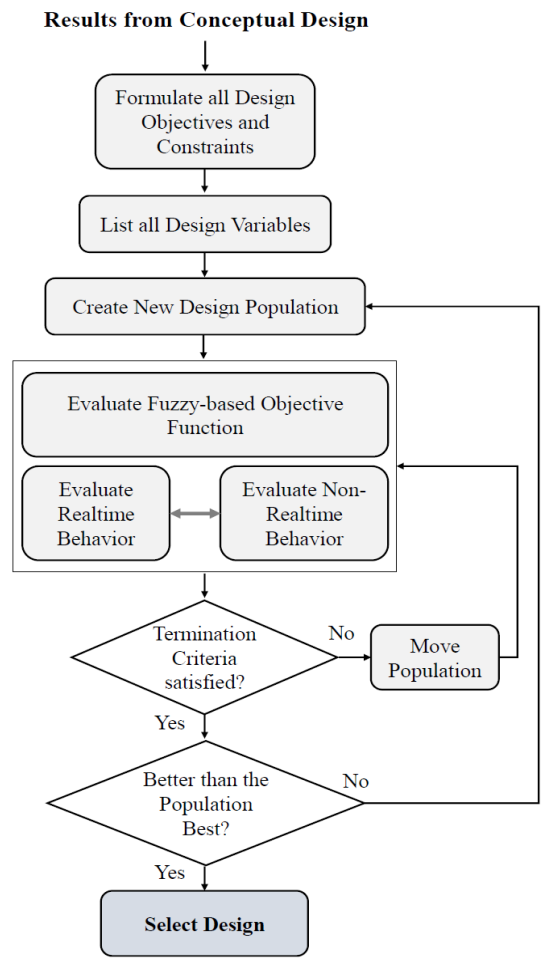

Fig. 3: Proposed detailed design procedure

## 3. Fuzzy Decision Support and Aggregation

### 3.1. Criteria Aggregation

The problem of aggregating criteria functions to form overall decision functions is of considerable importance in many disciplines. A primary factor in the determination of the structure of such aggregation functions is the relationship between the criteria involved. Choquet integral is a nonlinear fuzzy integral that has been successfully used for the aggregation of criteria in the presence of interactions. For mechatronics design and after quantifying all MMP elements and corresponding subsets, an effective comparison algorithm is needed. A global concept score (GCS) as a multi-criteria evaluation index can be defined to enable the designers to compare the feasible generated design concepts. GCS can be expressed as follows:

$$GCS = S(m_1^*, m_2^*, \ldots, m_n^*). \prod_{i=1}^{m} g(m_i), \qquad (3)$$

where $m_i^*$ are the normalized criteria values, $S(.)$ represents an aggregation function which, in this paper, is the Choquet integral, and $g(m_i)$ indicates whether a design constraint has been met (binary value).

### 3.2. Fuzzy Measures and Choquet Integrals

Choquet integral provides a weighting factor for each criterion, and also for each subset of criteria. Using Choquet integrals is a very effective way to measure an expected utility when dealing with uncertainty, which is the case in design in general and mechatronics design in particular. The main advantage of using this technique over other methods, such as weighted mean, is that by defining a weighting factor for each subset of criteria, the interactions between multiple objectives and criteria can be easily taken into account as well as their individual importance. To help a better understanding of the proposed solution, we will state some definitions in the following paragraphs.

**Definition 1:** The weighting factor of a subset of criteria is represented by a fuzzy measure on the universe $N$ satisfying the following fuzzy measure ($\mu$) equations:

$$\mu(\phi) = 0, \qquad \mu(N) = 1. \qquad (4)$$

$$A \subseteq B \subseteq N \ \rightarrow \mu(A) \leq \mu(B). \qquad (5)$$

where A and B represent the fuzzy sets (Sugeno 1975). Equation (4) represents the boundary conditions for fuzzy measures while Equation (5) is also called the monotonicity property of fuzzy measures.

**Definition 2:** Let $\mu$ be a fuzzy measure on vector $X$, whose $n$ elements are denoted by $x_1, x_2, \ldots, x_n$. The discrete Choquet integral of a function $f: X \rightarrow \mathbb{R}^+$ with respect to $\mu$ is defined by:

$$C_\mu(f) = \sum_{i=1}^{n}(f(x_i) - f(x_{i-1}))\mu(A_{(i)}), \tag{6}$$

where indices have been permuted so that $0 \leq f(x_1) \leq f(x_2) \leq \cdots \leq f(x_n)$ and $A_{(i)} = \{(i), \ldots (n)\}$, and $A_{(n+1)} = \emptyset$ while $f(x_0) = 0$. Figure 4 gives a graphical illustration of Choquet integral compared to a weighted sum while Table 1 shows the most common semantic interactions among criteria pairs and the corresponding fuzzy measures.

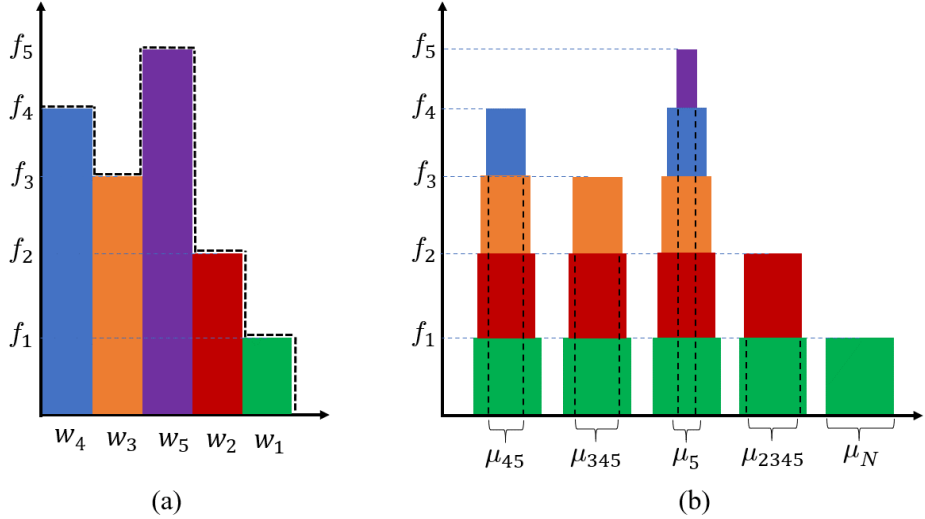

Fig. 4: Graphical illustration of (a) Weighted Sum and (b) Choquet integral

Table 1- Fuzzy Interactions and Measurements

| # | Description of Interaction | Fuzzy Measurement |
|---|---|---|
| I | Negative Correlation | $\mu(i,j) > \mu(i) + \mu(j)$ |
| II | Positive Correlation | $\mu(i,j) < \mu(i) + \mu(j)$ |
| III | Substitution | $\mu(T) \underset{T \subseteq Y \setminus i,j}{<} \begin{Bmatrix} \mu(T \cup i) \\ \mu(T \cup j) \end{Bmatrix} \approx \mu(T \cup i \cup j)$ |
| IV | Veto Effect | $\mu(T) \approx 0$ if $T \subset Y, i \notin T$ |
| V | Pass Effect | $\mu(T) \approx 1$ if $T \subset Y, i \in T$ |
| VI | Complementarity | $\mu(T) \underset{T \subseteq Y \setminus i,j}{=} \begin{Bmatrix} \mu(T \cup i) \\ \mu(T \cup j) \end{Bmatrix} < \mu(T \cup i \cup j)$ |

A lattice representation can be used for describing fuzzy measures in the case of a finite number of criteria. Figure 4 gives an illustration when $n = 4$. Please note that for simplicity we use $\mu_{ij}$ instead of $\mu(\{i,j\})$.

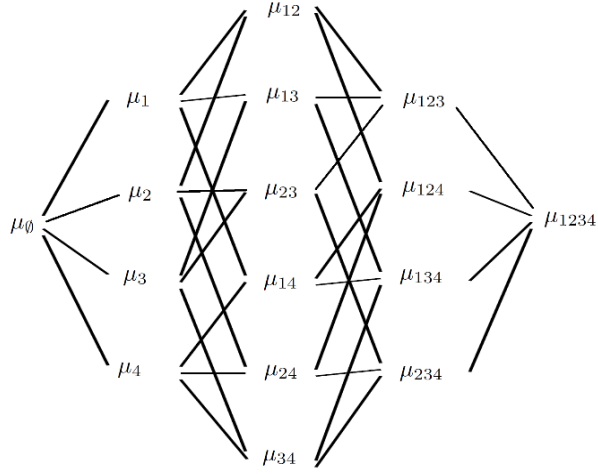

Fig. 5: Lattice of the coefficients of a fuzzy measure (*n=4*)

**Definition 3:** Let $\mu$ be a fuzzy measure. The interaction index $I(\mu, ij)$ for any pair of criteria $i$ and $j$ is defined as follows (Marichal 2002):

$$I(\mu, ij) = \sum_{T \subseteq N \setminus i,j} \frac{(n-t-2)!\,t!}{(n-1)!} [\mu(T \cup ij) - \mu(T \cup i) - \mu(T \cup j) + \mu(T)]. \tag{7}$$

where $T$ is a subset of criteria. The interaction index ranges in [-1, 1].

**Definition 4:** The importance index $\phi(\mu, i)$ for a criterion $i$ is computed by the Shapley value ($\phi$) (Marichal 2002), which is defined as:

$$\phi(\mu, i) = \sum_{T \subseteq N \setminus i} \frac{(n-t-1)!\,t!}{n!} [\mu(T \cup i) - \mu(T)]. \tag{8}$$

The Shapley value ranges between [0, 1] and represents a true sharing of the total amount $\mu(N)$, since:

$$\sum_{i=1}^{n} \phi(\mu, i) = \mu(N) = 1. \tag{9}$$

It is convenient to scale these values by a factor $n$, so that an importance index greater than 1 indicates an attribute more important than the average.

**Lemma:** If the coefficients $\mu(\{i\})$ and $\mu(\{i,j\})$ are given for all $i, j \in N$, then the necessary and sufficient conditions that $\mu$ is a 2-additive measure are:

$$\sum_{\{i,j\} \subseteq N} \mu(\{i,j\}) - (n-2) \sum_{i \in N} \mu(\{i\}) = 1 \qquad \text{(Normality)} \tag{10}$$

$$\mu(\{i\}) \geq 0, \forall i \in N \qquad \text{(Non-negativity)} \tag{11}$$

$$\forall A \subseteq N, |A| \geq 2, \forall k \in A,$$
$$\sum_{i \in A\setminus\{k\}} (\mu(\{i,k\}) - \mu(\{i\})) \geq (|A| - 2)\mu(\{k\}) \qquad \text{(Monotonicity)} \qquad (12)$$

The expression of the *2-additive* Choquet is:

$$C_\mu(f) = \sum_{i=1}^{n} \phi(\mu, i) f(x_i) - \frac{1}{2} \sum_{\{i,j\} \subseteq N} I(\mu, ij) |f(x_i) - f(x_j)| \qquad (13)$$

Here, $I(\mu, ij) = 0$ means criteria $i$ and $j$ are independent while $I(\mu, ij) > 0$ means there is a complementary among $i$ and $j$ and that for the decision-maker, both criteria have to be satisfactory in order to get a satisfactory alternative. If $I(\mu, ij) < 0$ then there is substitutability or redundancy among $i$ and $j$. This means that for the decision-maker, the satisfaction of one of the two criteria is sufficient to have a satisfactory alternative. It is worthy to note that a *positive correlation* leads to a *negative interaction index*, and vice versa. The fuzzy measures should be specified in such a way that the desired overall importance and the interaction indices are satisfied.

### 3.3. Fuzzy-based Design Schemes

Using the formulations described in 3.1 and 3.2 for aggregation of interacting criteria, the procedure of conceptual and detailed design can now be illustrated as Figure 4a and 4b. In the conceptual stage and using the assessed MMP, the fuzzy measures are used to specify the weight of importance and interactions amongst design criteria. Then, each design alternative is evaluated by incorporating a Choquet aggregation function and a rank on the elit set of concepts is provided to port to the detailed design stage. In detailed design, a multi-objective optimization process is considered to concurrently design for realtime and non-realtime variables that correspond to the optimal behavior of the overall system. In order to provide the optimization algorithm with an interactive objective function that includes all the design requirements from various disciplines, a cascade Choquet integral-based aggregation is used. This takes into account all the interactions amongst design objectives and also their relative importance in the design process.

## 4. Identification of Fuzzy Measures

As shown in Figure 4, in both stages, identification of fuzzy measures is a crucial stage that should be carefully done to correctly reflect on the decision making process. We now address the problem of identification of $(2^n - 2)$ fuzzy measures, $\mu$, taking into account the monotonicity relations between the coefficients and the preferences specified by requirements and the decision-makers. Four different approaches are essentially discussed here;

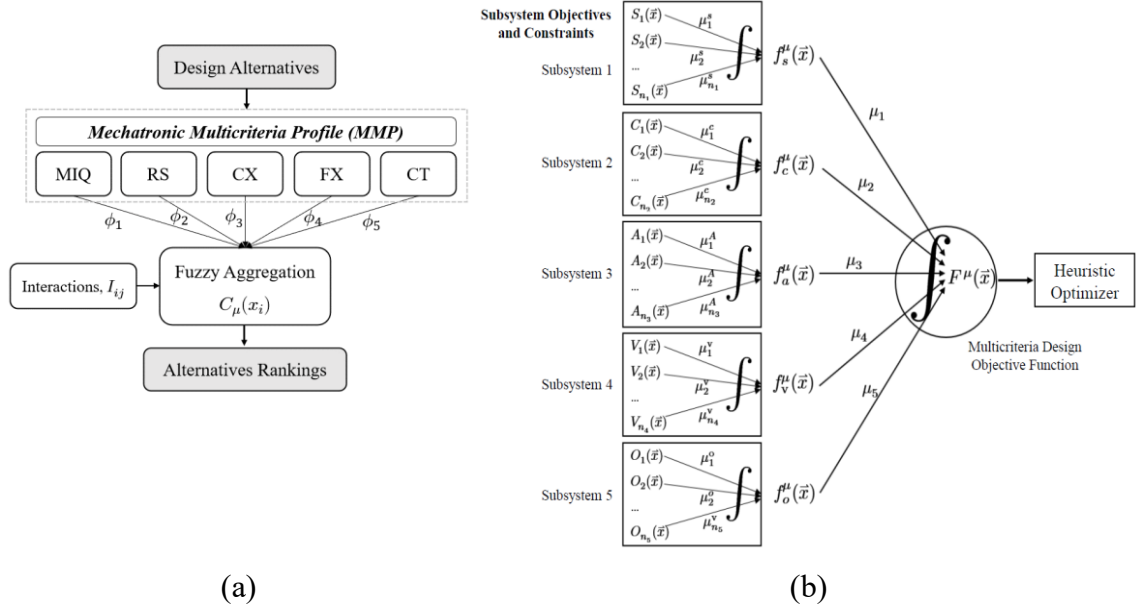

(a)                                           (b)

Fig. 6: Fuzzy based design of a mechatronic system for (a) concept evaluation, and (b) detailed design using a multiobjective optimization scheme

### *4.1. Identification using Sugeno measures*

As the number of criteria, $n$, grows specifications of the fuzzy measures using the aforementioned methods become more and more difficult. (Sugeno 1975) created a way to automatically generate the entire lattice based on just the singleton $\mu_i$ densities, thus $(2^n - 2 - n)$ values. The Sugeno $\lambda$-fuzzy measure has the following additional property: If $A, B \in \Omega$ and $A \cap B = \emptyset$,

$$\mu(A \cup B) = \mu(A) + \mu(B) + \lambda\mu(A)\mu(B). \tag{14}$$

It is proven that a unique $\lambda$ can be found by solving the following equation:

$$\lambda + 1 = \prod_{i=1}^{n}(1 + \lambda\mu_i), \quad -1 < \lambda < \infty, \lambda \neq 0 \tag{15}$$

where $\mu_i = \mu\{x_i\}$. Thus, the $n$ densities determine the $2^n$ values of a Sugeno measure. There are three cases with regards to the singleton measures;

$$\text{If } \sum_{i=1}^{n} \mu_i > \mu(N) \text{ then, } -1 < \lambda < \infty. \tag{16}$$

$$\text{If } \sum_{i=1}^{n} \mu_i = \mu(N) \text{ then, } \lambda = 0. \tag{17}$$

$$\text{If } \sum_{i=1}^{n} \mu_i < \mu(N) \text{ then, } \lambda > 0. \tag{18}$$

The process of using the Sugeno method to identify the full lattice of fuzzy measures is summarized in Figure 7.

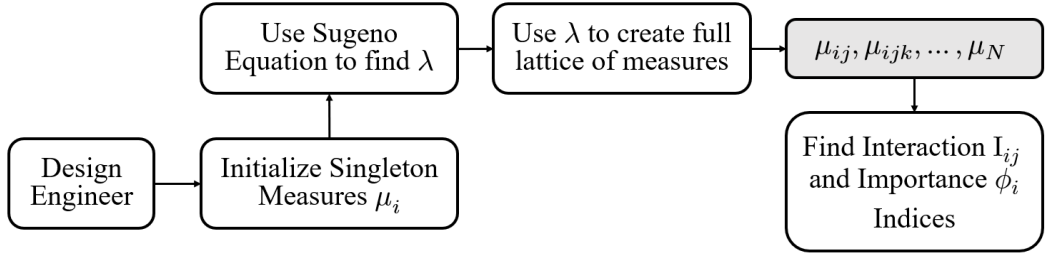

Fig. 7: Identification of fuzzy measures using Sugeno process

### *4.2. Identification based on learning data*

Having a set of learning data in hand, the parameters of a Choquet integral model can be identified by minimizing an error criterion. Suppose that $(f_k, y_k)$, $k = 1, 2, \ldots, l$ are learning data where $f_k = [f^k(x_1), \ldots, f^k(x_n)]^T$ is an $n$-dimensional input vector, containing the degrees of satisfaction or quantified assessment values of alternative (concept) $k$ with respect to criteria 1 to $n$, and $y_k$ is the global evaluation of object $k$ (not necessarily an aggregated value). There must be at least $l = \frac{n!}{\left[\left(\frac{n}{2}\right)!\right]^2}$ (when $n$ is even) or $l = \frac{n!}{\left[\frac{n-1}{2}\right]!\left[\frac{n+1}{2}\right]!}$ (when $n$ is odd) sets of learning data (Grabisch, Nguyen et al. 2013). Then, one can try to identify the best fuzzy measure $\mu^*$ so that the squared error criterion (E) is minimized (Grabisch 1996).

$$E^2 = \sum_{k=1}^{l} \left[ C_\mu\left(f^k(x_1), \ldots, f^k(x_n)\right) - y_k \right]^2 \qquad (19)$$

Under a quadratic program form, we have:

$$\min\left( E^2 = \left(\frac{1}{2}\mathbf{u}^t \mathbf{D} \mathbf{u} + \mathbf{c}^t \mathbf{u}\right) \right) \qquad (20)$$

where $\mathbf{u}$ is a $(2^n - 2)$ dimensional vector containing all the coefficients of the fuzzy measure $\mu$, except for $\mu_\emptyset = 0$ and $\mu_N = 1$, as follows:

$$\mathbf{u} = \left[ [\mu_i], [\mu_{ij}], [\mu_{ijk}], [\mu_{ijkl}], \ldots \right]^T \qquad (21)$$

It is important to note that the components of $\mathbf{u}$ are not independent of each other because fuzzy measures must satisfy a set of monotonicity relations. Moreover, $\mathbf{D}$ is a symmetric $(2^n - 2)$ dimensional matrix and $\mathbf{c}$ is a $(2^n - 2)$ dimensional vector. The first set of constraints contains the measures monotonicity constraints described as follows:

$$\mathbf{Au} + \mathbf{b} \leq 0 \qquad (22)$$

where matrix $\mathbf{A}$ is a $n(2^{n-1} - 1) \times (2^n - 2)$ dimensional matrix and $\mathbf{b}$ is a $n(2^{n-1} - 1)$ vector defined by:

$$\mathbf{b} = [0, \ldots, 0, \underbrace{-1, \ldots, -1}_{n}]^T. \tag{23}$$

More precisely for Equation (18) we have:

$$C_\mu(f_k) = \mathbf{c}_k^t \cdot \mathbf{u} + f^k(x_1), \tag{24}$$

where $\mathbf{c}_k$ is a $(2^n - 2)$ dimensional vector containing the differences $f(x_i) - f(x_{i-1})$, $i = 2, \ldots, n$, so that there are at most $(n - 1)$ non-zero terms in it, which are all positive. Accordingly, we attain:

$$\mathbf{c} = 2 \sum_{k=1}^{l} (f^k(x_1) - y_k) \mathbf{c}_k. \tag{25}$$

Additionally, $D_k$ is a $(2^n - 2)$ dimensional square matrix where:

$$\mathbf{D} = 2 \sum_{k=1}^{l} \mathbf{D}_k = 2 \sum_{k=1}^{l} \mathbf{c}_k \mathbf{c}_k^T. \tag{26}$$

Thus, we can rewrite the program in Equation (24) as:

$$\min \left( E^2 = 2 \sum_{k=1}^{l} \mathbf{u}^T \mathbf{c}_k \mathbf{c}_k^T \mathbf{u} + 2 \sum_{k=1}^{l} \mathbf{c}_k^T \cdot \mathbf{u} \, (f^k(x_1) - y_k) \right) \tag{27}$$

$$\text{Subj. to } \mathbf{A}\mathbf{u} + \mathbf{b} \leq 0$$

Since $u^T \mathbf{D} u$ consists of a sum of squares, thus for all $u \geq 0$, $u^T \mathbf{D} u \geq 0$ and $\mathbf{D}$ is positive semidefinite. The above quadratic program has a unique (global) minimum since the criterion to be minimized is convex. This solution can be a point or a convex set in $[0, 1]^{2n-2}$. This program can be solved by any standard method of quadratic optimization, although matrix $\mathbf{D}$ may be ill-conditioned ($rank < 2^n - 2$) since based on the definition of vector $\mathbf{c}_k$, matrix $\mathbf{D}$ contains columns and rows of zeroes. Obviously, this effect will disappear if the number of training data increases.

Now, we can take into account the decision maker's (DM) preferences with regards to the importance of criteria and interactions among criterion pairs as constraint relations;

$$\mu(A \cup i) - \mu(A) \geq 0, \quad \forall i \in N, \forall A \in N \setminus i \tag{28}$$

$$C_\mu(f) - C_\mu(\hat{f}) \geq \delta_C \tag{29}$$

$$\phi(\mu, i) - \phi(\mu, j) \geq \delta_\phi \tag{30}$$

$$\text{Constraints on } I(\mu, ij) \tag{31}$$

The process of using learning data in addition to the designer's preferences to identify the fuzzy measures is summarized in Figure 8.

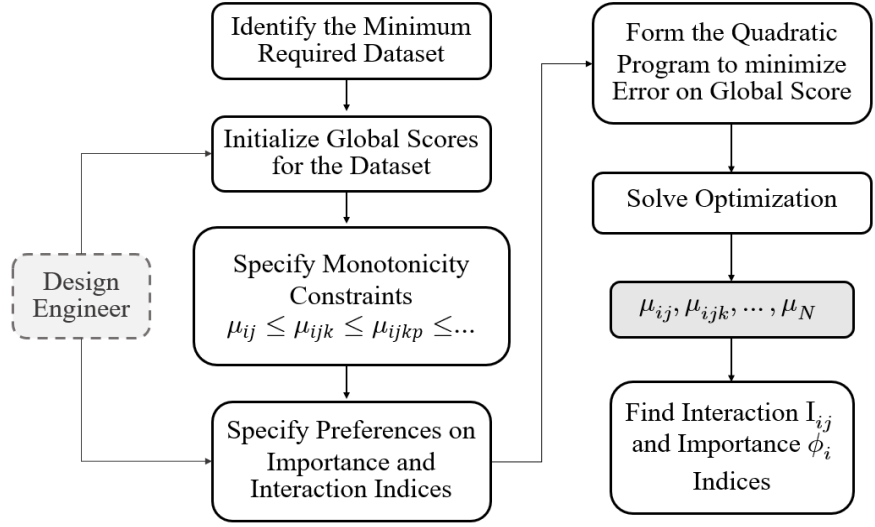

Fig. 8: Identification of fuzzy measures using learning data and quadratic programing

*4.3. Identification based on fuzzy measure semantics and learning data*

To reduce the complexity and provide better guidelines for the identification of measures, the combination of semantical considerations with learning data can lead to a more efficient algorithm. With this approach, the objective would be to minimize the distance to the additive equidistributed fuzzy measure defined by $\mu_j = 1/n$. Consequently, instead of trying to minimize the sum of the squared errors between model output and data, we try to minimize the distance to the additive equidistributed measure set $\mathbf{u_0}$. Thus, we can have the following quadratic form:

$$\text{Min } J = \tfrac{1}{2}(u - u_0)^T(u - u_0) \tag{32}$$

$$\text{Subj. to } \mathbf{Au} + \mathbf{b} \leq 0$$

Here, training data are no longer in the objective function, but are used as the second set of constraints;

$$y_k - \delta_k \leq \mathbf{c}_k^t \cdot \mathbf{u} + f(x_1) \leq y_k + \delta_k \tag{33}$$

Moreover, the decision-maker needs to express some preferences as the relative importance of the criteria and on their mutual interactions, such that:

$$\mu(A) \leq \eta\mu(B) \tag{34}$$

$$\mu(A \cup B) = \mu(A) + \lambda\mu(B) \tag{35}$$

where $\mu_A \geq \mu_B$ and $\eta$ defines the degree of the relative importance of A with respect to B. For the interactions between criteria $A$ and $B$, $\lambda \in [0,1]$ and $A$ and $B$ are fully dependent when $\lambda = 0$, and independent when $\lambda = 1$. Support (synergy) between A and B can be modeled by:

$$\mu(A \cup B) = \mu(A) + \mu(B) + \gamma(1 - \mu(A) - \mu(B)) \tag{36}$$

where $\gamma$ specifies the level of support between criteria pairs. All these constraints based on the decision-maker's preferences can be used to modify the initial monotonicity constraint by adding to the initial **A** and **b** and form a new constraint as:

$$\mathbf{A'u + b'} \leq 0 \tag{37}$$

## 5. Case Study: Conceptual Design of a Vision-Guided Quadrotor Drone

Recently, the quadrotors are being deployed as highly maneuverable aerial robots which have the ability of easy hover, take off, fly, and land in small and remote areas (Mohebbi, Achiche et al. 2015). Recent technological advances in energy storage devices, sensors, actuators, and information processing have boosted the development of Unmanned Aerial Vehicle (UAV) platforms with significant capabilities. Unmanned Quadrotor Helicopters (UQH) are excellent examples of highly coupled mechatronic systems where the disciplines of aerodynamics, structures, and materials, flight mechanics, and control are acting upon each other in a typical flight condition. Moreover, the integration of vision sensors with robots has helped solve the limitation of operating in non-structured environments (Mohebbi, Keshmiri et al. 2016).

Here, the discussed fuzzy measure identification methods are utilized in a conceptual design process using the multicriteria mechatronic profile (MMP) for a vision-guided quadrotor UAV. From our previous work (Mohebbi, Achiche et al. 2016), we have chosen four concepts to study the proposed design method. Table 2 shows the design alternative and the corresponding sub-systems and components. Based on the material used, the frame structure and subsystems selected for one specific concept, the total mass, required power, payload, maximum allowable inertia moment, force, and bandwidth can be also easily estimated. An approximation of the total cost can also be calculated based on the components and manufacturing process. Table 3 briefly gives the results for the estimated values for the proposed concepts.

Table 2: Design alternatives adopted from (Mohebbi, Achiche et al. 2016)

|  | **Concept I** | **Concept II** | **Concept III** | **Concept IV** |
|---|---|---|---|---|
| Frame Structure | X-shape | H-shape | X-shape | H-shape |
| Material | AL. | AL. | Poly. | Poly. |
| Motors | Brushed DC | Brushed DC | Brushless DC | Brushless AC |
| Motor Encoder | Optical | Magnetic | Optical | Magnetic |
| Visual Servo. | PBVS | PBVS | IBVS | IBVS |
| Camera Config. | Mono | Stereo | Stereo | Mono |
| Motion Control. | PID | LQR | PID | LQR |
| Position Sensor | GPS +Accel. | Motion Cam | GPS +Accel. | GPS +Accel. |
| Battery | Li-ion | Li-Poly. | Li-ion | Li-Poly. |

Table 3: Estimated design parameters for generated concepts

|  | Concept I | Concept II | Concept III | Concept IV |
|---|---|---|---|---|
| Power (W) | 450 | 500 | 350 | 400 |
| Max Inertia Moment (kg.m$^2$) | 5E-3 | 5.2E-3 | 4E-3 | 4.5E-3 |
| Bandwidth (Hz) | 70 | 70 | 60 | 60 |
| Payload (Kg) | 0.5 | 0.5 | 0.6 | 0.6 |
| Cost (unit) (normal.) | 0.8 | 1 | 0.7 | 0.7 |

Ultimately, by using a set of intuitive Choquet fuzzy measures the evaluations for all concepts and corresponding design criteria are listed in Table 4. Details of the criteria assessment and calculations are thoroughly discussed and exemplified in our previous work introducing the Mechatronic Multicriteria Profile (Mohebbi, Achiche et al. 2014, Mohebbi, Achiche et al. 2016). The fuzzy measures used in the previous study were obtained intuitively by the authors and a group of 30 researchers (all specialized in system design and mechatronics) through a questionnaire. In this questionnaire, the participants were asked to reflect their intuitive idea about the importance of each criterion in designing a good mechatronic product in terms of a score between 1 and 10. Moreover, the degree of correlation between each pair of criteria or the effect of increasing criterion $i$ on criterion $j$ were also asked and reflected in terms of a score between -10 and 10. Then, the obtained values were transformed into fuzzy measures that fit the requirements discussed in Equations (4, 5, 11-13). These measures are shown in Table 5.

Table 4: Concept Evaluations for design alternatives (Mohebbi, Achiche et al. 2014, Mohebbi, Achiche et al. 2016).

| *MMP* | *Concept I* | *Concept II* | *Concept III* | *Concept IV* |
|---|---|---|---|---|
| MIQ | 0.84 | 0.84 | 1 | 1 |
| RS | 0.86 | 0.91 | 0.93 | 1 |
| CX | 0.85 | 0.69 | 0.93 | 0.89 |
| FX | 0.91 | 0.96 | 0.91 | 0.88 |
| CT | 1 | 0.78 | 0.94 | 0.91 |
| **$GCS_\mu$** | 0.89 | 0.83 | 0.96 | 0.94 |

$$\Phi = [\phi_1, \phi_2, \phi_3, \phi_4, \phi_5] = [0.2085, 0.2612, 0.1598, 0.1431, 0.2020]. \quad (38)$$

We remind that in order to calculate a Choquet integral and its corresponding measures, a permutation on the criteria values should be initially performed in such a way that $0 \leq f(x_1) \leq f(x_2) \leq \cdots \leq f(x_n)$. Although, throughout our case study and to avoid any confusion, we reshape the outputs for measures and also importance indices at the end of the identification algorithm so that the following order always persists:

$$\Phi = [\phi_1, \phi_2, \phi_3, \phi_4, \phi_5] = [\phi_{MIQ}, \phi_{RS}, \phi_{CX}, \phi_{FX}, \phi_{CT}]. \quad (39)$$

Table 5: Fuzzy measures for the conceptual design of a Quadrotor drone equipped with a visual servoing system

| $\mu_1$ = 0.23 | $\mu_{12}$ = 0.45 | $\mu_{13}$ = 0.47 | $\mu_{14}$ = 0.34 | $\mu_{15}$ = 0.51 |
|---|---|---|---|---|
| $\mu_{123}$ = 0.61 | $\mu_2$ = 0.29 | $\mu_{23}$ = 0.52 | $\mu_{24}$ = 0.42 | $\mu_{25}$ = 0.56 |
| $\mu_{124}$ = 0.60 | $\mu_{135}$ = 0.69 | $\mu_3$ = 0.17 | $\mu_{34}$ = 0.35 | $\mu_{35}$ = 0.33 |
| $\mu_{125}$ = 0.67 | $\mu_{145}$ = 0.67 | $\mu_{245}$ = 0.73 | $\mu_4$ = 0.16 | $\mu_{45}$ = 0.41 |
| $\mu_{134}$ = 0.63 | $\mu_{234}$ = 0.68 | $\mu_{345}$ = 0.49 | $\mu_{235}$ = 0.62 | $\mu_5$ = 0.22 |
| $\mu_{1234}$ = 0.77 | $\mu_{1235}$ = 0.84 | $\mu_{1345}$ = 0.84 | $\mu_{2345}$ = 0.78 | $\mu_{1245}$ = 0.82 |

### *5.1. Identification using Sugeno measures*

Based on Equations (36-37) for five criteria illustrated in Table 4 we have:

$$\lambda + 1 = (\lambda\mu_1 + 1)(\lambda\mu_2 + 1)(\lambda\mu_3 + 1)(\lambda\mu_4 + 1)(\lambda\mu_5 + 1) \qquad (40)$$
$$-1 < \lambda < \infty, \qquad \lambda \neq 0.$$

where for $\mu_i$ we use the values from Table 5. The solution of the above equation yields $\lambda = 0.0255$ and consequently, we attain the results for fuzzy measures listed in Table 6.

Table 6- Fuzzy measures identified using Sugeno $\lambda$- measures

| $\mu_1$ = 0.22 | $\mu_{12}$ = 0.4613 | $\mu_{13}$ = 0.3910 | $\mu_{14}$ = 0.3809 | $\mu_{15}$ = 0.4211 |
|---|---|---|---|---|
| $\mu_{123}$ = 0.6333 | $\mu_2$ = 0.24 | $\mu_{23}$ = 0.4110 | $\mu_{24}$ = 0.4010 | $\mu_{25}$ = 0.4412 |
| $\mu_{124}$ = 0.6232 | $\mu_{135}$ = 0.5930 | $\mu_3$ = 0.17 | $\mu_{34}$ = 0.3307 | $\mu_{35}$ = 0.3709 |
| $\mu_{125}$ = 0.6637 | $\mu_{145}$ = 0.5828 | $\mu_{245}$ = 0.6030 | $\mu_4$ = 0.16 | $\mu_{45}$ = 0.3608 |
| $\mu_{134}$ = 0.5526 | $\mu_{234}$ = 0.5727 | $\mu_{345}$ = 0.5324 | $\mu_{235}$ = 0.6131 | $\mu_5$ = 0.20 |
| $\mu_{1234}$ = 0.7959 | $\mu_{1235}$ = 0.8366 | $\mu_{1345}$ = 0.7554 | $\mu_{2345}$ = 0.7756 | $\mu_{1245}$ = 0.8264 |

The fuzzy measures obtained by the Sugeno $\lambda-$ method yield the following importance indices:

$$\Phi = [\phi_1, \phi_2, \phi_3, \phi_4, \phi_5] = [0.2221, 0.2422, 0.1718, 0.1617, 0.2020]. \qquad (41)$$

### *5.2. Identification using a learning set*

As mentioned before, in order to identify the fuzzy measures, it is possible to employ a "learning set"—a number of objects whose assessment is manually performed by the decision-maker (DM). According to (Grabisch, Nguyen et al. 2013), the minimum

number of data set we need to solve the squared error minimization program (19) is equal to:

$$l = \frac{n!}{\left[\frac{n-1}{2}\right]! \left[\frac{n+1}{2}\right]!} = \frac{5!}{\left[\frac{5-1}{2}\right]! \left[\frac{5+1}{2}\right]!} = 10. \tag{42}$$

Accordingly, we need to provide 10 sets of criteria evaluation and corresponding global concept scores. The vector of variables contains the 30 fuzzy measures and as for the monotonicity constraints described in Equation (21) we have the following matrices:

$$\mathbf{A}_{[75 \times 30]}, \mathbf{u}_{[30 \times 1]}, \mathbf{b} = \left[0, \dots, 0, \underbrace{-1, \dots, -1}_{5}\right]^T_{[75 \times 1]} \tag{43}$$

in which we describe all 75 monotonicity relations such as:

$$\begin{aligned}
\mu_1 &\leq \mu_{12}, \dots, \mu_5 \leq \mu_{45}, \\
\mu_{12} &\leq \mu_{123}, \dots, \mu_{45} \leq \mu_{345}, \\
\mu_{123} &\leq \mu_{1234}, \dots, \mu_{345} \leq \mu_{2345}, \\
\mu_{1234} &\leq 1, \dots, \mu_{2345} \leq 1.
\end{aligned} \tag{44}$$

In order to form the objective function from Equation (19) we also need to form the matrix $\mathbf{D}$ and vector $\mathbf{c}$ which have the following format:

$$\mathbf{D}_{[30 \times 30]}, \mathbf{c}_{[30 \times 1]}, \mathbf{c}_{k[30 \times 1]}$$

$$\mathbf{c} = 2 \sum_{k=1}^{10} (f^k(x_1) - y_k) \mathbf{c}_k, \tag{45}$$

$$\mathbf{D} = 2 \sum_{k=1}^{10} \mathbf{D}_k = 2 \sum_{k=1}^{10} \mathbf{c}_k \mathbf{c}_k^T. \tag{46}$$

In which $\mathbf{c}_k$ is a 30- dimensional vector containing the differences $f(x_i) - f(x_{i-1})$, $i = 2, \dots, 5$ so that there are at most 4 non-zero terms in it, which are all positive. Consequently, we get:

$$\begin{aligned}
\mathbf{c}_k(5) &= f^k(x_5) - f^k(x_4), \\
\mathbf{c}_k(15) &= f^k(x_4) - f^k(x_3), \\
\mathbf{c}_k(25) &= f^k(x_3) - f^k(x_2), \\
\mathbf{c}_k(30) &= f^k(x_2) - f^k(x_1), \\
\mathbf{c}_k(i) &= 0, \qquad (\forall i \neq 5, 15, 25, 30)
\end{aligned} \tag{47}$$

Finally, the decision-maker's preferences can be taken into account using the constraints listed in Table 7.

Table 7 – Decision-maker's preferences on criteria relations

| Maximum separation of alternatives |
|---|
| $C_\mu(f) - C_\mu(f') \geq \delta_c \ (\delta_c = 0.05)$ |
| Preferences on the importance of criteria |

| $\phi_2 - \phi_1 \geq \epsilon$ | $\phi_2 - \phi_5 \geq \epsilon$ |
|---|---|
| $\phi_1 - \phi_3 \geq \epsilon$ | $\phi_5 - \phi_3 \geq \epsilon$ |
| $\phi_1 - \phi_4 \geq \epsilon$ | $\phi_5 - \phi_4 \geq \epsilon$ |
| $\phi_2 - \phi_3 \geq \epsilon$ | $\phi_1 = \phi_5$ |
| $\phi_2 - \phi_4 \geq \epsilon$ | $\phi_3 = \phi_4$ |
| Preferences on the interactions between criteria pairs ||
| $I(1,5) - I(1,3) \geq \epsilon$ | $I(4,5) - I(3,4) \geq \epsilon$ |
| $I(2,5) - I(2,3) \geq \epsilon$ | $I(2,4) = I(3,4)$ |
| $I(1,3) - I(2,4) \geq \epsilon$ | $I(1,4) = I(3,5)$ |

The above problem will be solved here using MATLAB quadratic programming from the optimization toolbox and the method of "interior-point-convex". Table 8 shows the resulting values for the fuzzy measures.

Table 8 – Results for fuzzy measures identified using a learning set

| $\mu_1$ = 0.3292 | $\mu_{12}$ = 0.4502 | $\mu_{13}$ = 0.6366 | $\mu_{14}$ = 0.2985 | $\mu_{15}$ = 0.6416 |
|---|---|---|---|---|
| $\mu_{123}$ = 0.7983 | $\mu_2$ = 0.2829 | $\mu_{23}$ = 0.5137 | $\mu_{24}$ = 0.5615 | $\mu_{25}$ = 0.5332 |
| $\mu_{124}$ = 0.4398 | $\mu_{135}$ = 0.7296 | $\mu_3$ = 0.1901 | $\mu_{34}$ = 0.4698 | $\mu_{35}$ = 0.1789 |
| $\mu_{125}$ = 0.8048 | $\mu_{145}$ = 0.6610 | $\mu_{245}$ = 0.8620 | $\mu_4$ = 0.2584 | $\mu_{45}$ = 0.5167 |
| $\mu_{134}$ = 0.6273 | $\mu_{234}$ = 0.8137 | $\mu_{345}$ = 0.5088 | $\mu_{235}$ = 0.5446 | $\mu_5$ = 0.2082 |
| $\mu_{1234}$ = 0.8093 | $\mu_{1235}$ = 0.9334 | $\mu_{1345}$ = 0.8093 | $\mu_{2345}$ = 0.8093 | $\mu_{1245}$ = 0.8444 |

The above results will lead to the following importance indices:
$$\Phi = [\phi_1, \phi_2, \phi_3, \phi_4, \phi_5] = [0.2145, 0.2535, 0.1701, 0.1597, 0.1967]. \qquad (48)$$

### *5.3. Identification based on fuzzy measure semantics and learning data*

In order to use Equations (33-35) for modeling the relations between criteria pairs, we define the proper linguistics as described in Tables 9-11.

Table 9 - Linguistic representation of the relative importance of criteria

| Relative Importance | Value |
|---|---|
| same level | $0.9 \leq \eta \leq 1.1$ |
| $A$ is a little more important than $B$ | $1.1 \leq \eta \leq 1.3$ |
| $A$ is more important than $B$ | $1.3 \leq \eta \leq 1.7$ |
| $A$ is quite more important than $B$ | $1.7 \leq \eta \leq 1.9$ |

Table 10 - Linguistic representation of dependence between criteria

| Criteria Dependence | Value |
|---|---|
| highly dependent | $\lambda = 0.0$ |
| Dependent | $0.0 \leq \lambda \leq 0.5$ |
| a little dependent | $0.5 \leq \lambda \leq 1.0$ |
| Independent | $\lambda = 1.0$ |

Table 11 - Linguistic representation of support between criteria

| Criteria Synergy | Value |
|---|---|
| high support | $\gamma = 1.0$ |
| Support | $0.5 \leq \gamma \leq 1.0$ |
| a little support | $0.0 \leq \gamma \leq 0.5$ |

These linguistics in addition to the monotonicity conditions are translated into the constraints as the decision-maker's preferences as described in Table 12.

Table 12 - Decision-maker's preferences as linear constraints

| Relative Importance of criteria | |
|---|---|
| $\mu_2 \leq 1.3\mu_1$ | $0.9\mu_4 \leq \mu_3 \leq 1.1\mu_4$ |
| $\mu_1 \leq 1.3\mu_4$ | $0.9\mu_3 \leq \mu_5 \leq 1.1\mu_3$ |
| $\mu_2 \leq 1.7\mu_4$ | $0.9\mu_5 \leq \mu_1 \leq 1.1\mu_5$ |
| $\mu_2 \leq 1.7\mu_3$ | $0.9\mu_5 \leq \mu_1 \leq 1.1\mu_5$ |
| Dependence between criteria pairs | |
| $\mu_2 + 0.5\mu_3 \leq \mu_{23} \leq \mu_2 + \mu_3$ | $\mu_3 + 0.8\mu_4 \leq \mu_{34} \leq \mu_3 + \mu_4$ |
| $\mu_2 + 0.5\mu_4 \leq \mu_{24} \leq \mu_2 + \mu_4$ | $\mu_4 + 0.5\mu_5 \leq \mu_{45} \leq \mu_4 + \mu_5$ |
| $\mu_2 + 0.5\mu_5 \leq \mu_{25} \leq \mu_2 + \mu_5$ | |
| The synergy between criteria pairs | |
| $\mu_1 + \mu_4 + 0.3(1 - \mu_1 - \mu_4) \leq \mu_{14} \leq \mu_1 + \mu_4 + 0.7(1 - \mu_1 - \mu_4)$ | |
| $\mu_3 + \mu_5 + 0.3(1 - \mu_3 - \mu_5) \leq \mu_{35} \leq \mu_3 + \mu_5 + 0.7(1 - \mu_3 - \mu_5)$ | |
| $\mu_1 + \mu_2 \leq \mu_{12} \leq \mu_1 + \mu_2 + 0.3(1 - \mu_1 - \mu_2)$ | |

This approach can also include an interactive dialogue between the DM and the fuzzy measure identifying system. Solutions are presented to the decision-maker, who can refine them by specifying or modifying the relative importance and interaction between criteria if he is not satisfied with the solution. As an example, here we use the concept evaluation data from our previous work. As for the additive equidistributed singleton fuzzy measures we have:

$$M_0 = [0.2 \quad 0.2 \quad 0.2 \quad 0.2 \quad 0.2], \quad (49)$$

Moreover, we use the 10 training data sets from the previous section to form the following second set of constraints based on Equation (32) with $\delta_k = 0.35$;

$$0.54 \leq c_1^T \mathbf{u} + 0.84 \leq 1.24, \quad 0.47 \leq c_6^T \mathbf{u} + 0.64 \leq 1.17,$$
$$0.48 \leq c_2^T \mathbf{u} + 0.69 \leq 1.18, \quad 0.19 \leq c_7^T \mathbf{u} + 0.45 \leq 0.89,$$
$$0.61 \leq c_3^T \mathbf{u} + 0.91 \leq 1.31, \quad 0.53 \leq c_8^T \mathbf{u} + 0.75 \leq 1.23, \quad (50)$$
$$0.59 \leq c_4^T \mathbf{u} + 0.88 \leq 1.29, \quad 0.58 \leq c_9^T \mathbf{u} + 0.85 \leq 1.28,$$
$$0.44 \leq c_5^T \mathbf{u} + 0.72 \leq 1.14, \quad 0.07 \leq c_{10}^T \mathbf{u} + 0.35 \leq 0.77.$$

where $c_k^T$ is a $[1 \times 30]$ vector and can be calculated from Eq. 49, while for **u** we have:

$$\mathbf{u}_{[30 \times 1]} = \left[ [\mu_i], [\mu_{ij}], [\mu_{ijk}], [\mu_{ijkl...}], ... \right]^T. \quad (51)$$

By combining all the constraints in Eq. 52, Table 13 and also the monotonicity constraints, we can formulate a new linear constraint as $A'u + b' \leq 0$ and solve the quadratic program in Eq. 31. Again, by using MATLAB quadratic programming and the interior-point-convex algorithm we attain the following results:

Table 13 - Fuzzy measures identified using a learning set and design semantics

| $\mu_1$ = 0.3243 | $\mu_{12}$ = 0.4860 | $\mu_{13}$ = 0.6160 | $\mu_{14}$ = 0.2441 | $\mu_{15}$ = 0.6318 |
|---|---|---|---|---|
| $\mu_{123}$ = 0.8198 | $\mu_2$ = 0.2615 | $\mu_{23}$ = 0.4741 | $\mu_{24}$ = 0.5514 | $\mu_{25}$ = 0.5090 |
| $\mu_{124}$ = 0.4258 | $\mu_{135}$ = 0.7174 | $\mu_3$ = 0.1705 | $\mu_{34}$ = 0.4618 | $\mu_{35}$ = 0.1748 |
| $\mu_{125}$ = 0.8307 | $\mu_{145}$ = 0.6068 | $\mu_{245}$ = 0.8540 | $\mu_4$ = 0.2700 | $\mu_{45}$ = 0.5354 |
| $\mu_{134}$ = 0.5572 | $\mu_{234}$ = 0.7854 | $\mu_{345}$ = 0.5212 | $\mu_{235}$ = 0.5446 | $\mu_5$ = 0.2104 |
| $\mu_{1234}$ = 0.7810 | $\mu_{1235}$ = 0.9584 | $\mu_{1345}$ = 0.7810 | $\mu_{2345}$ = 0.7810 | $\mu_{1245}$ = 0.8255 |

Accordingly, we get the following Shapley values:
$$\Phi = [\phi_1, \phi_2, \phi_3, \phi_4, \phi_5] = [0.2085, 0.2612, 0.1598, 0.1431, 0.2020]. \quad (52)$$

## 6. Discussion and Comparison

Figure 9 describes the evolution of the full lattice of the fuzzy measure identified using the three methods discussed in this paper. Sugeno measures are among the most widely used fuzzy measures (Tahani and Keller 1990). Using $\lambda$-measures is an abstract and efficient way when there is not enough information about decision-maker preferences or the order of preference on alternatives or interaction and importance indices. It can rapidly generate the entire lattice of fuzzy measures based on just the singleton densities. Although, not all expert reasoning can be described by these measures and guessing the $\mu_i$ values intuitively is not a trivial process. In that case, this method can be also regarded as an optimization problem with all the preferences as constraints. Further information can be found in (Lee and LeeKwang 1995) since a complete identification process on $\lambda$-measures was not in the scope of this paper.

The identification based on learning data which uses minimization of the squared error needs only a global score, which can be provided by a ranking of the acts through a suitable mechanism. Besides the fuzzy measure, the output also provides an estimation of the model error. One important advantage of using this method is that having a proper optimization solver, it always provides a solution, which fits the given global scores. Moreover, the method does not need any information on the decision strategy (importance and interaction). It is perfectly suitable for identifying hidden decision behavior. Although it may temper with the concept rankings provided by the decision-maker.

In the identification based on combined fuzzy semantics and learning data, we need a ranking of the acts, not necessarily the global scores, a ranking on the importance of the criteria, and possibly some information on the interactions. There is no notion of model error in this approach in the sense that either there is a solution satisfying the constraints, or there is not. This method only requires a piece of ordinal information on the alternative and more importantly does not violate the ranking provided by the decision-maker. Although, the method ideally needs some information on the decision strategy. For example, one may use the method without any information on constraints but only the ranking of the relations. This makes the space of feasible solutions very big that the solution chosen may not have a real interpretation in terms of decision strategy. This method is more suitable when we need to define or build a decision strategy in terms of importance and interaction.

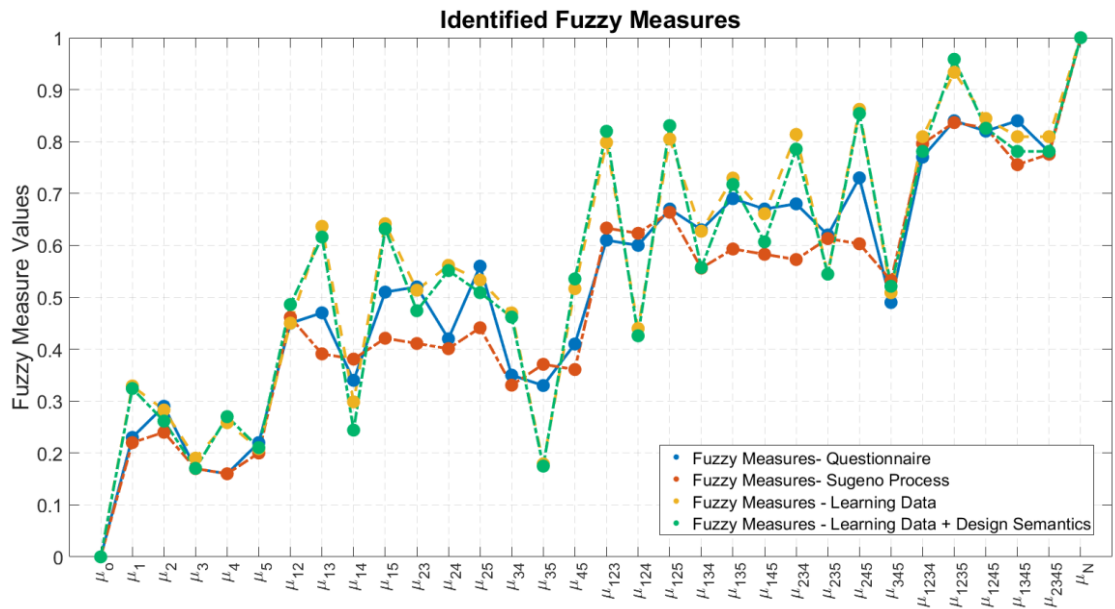

Fig. 9: Identification of fuzzy measures using learning data and quadratic programing

# 7. Conclusions

Mechatronic systems are seen as a combination of cooperative mechanical, electronics, and software components aided by various control strategies. They are often highly complex, because of the high number of their components, their multi-physical aspect, and the couplings between the different engineering domains involved which complexifies the design task. Therefore, to achieve a better design process as well as a better final product more efficiently, these couplings need to be considered in the early stages of the design process.

The concept of the Mechatronic Multicriteria Profile (MMP) has been previously introduced to facilitate fitting the intuitive requirements for decision-making in the presence of interacting criteria in conceptual design. The MMP includes five main elements: machine intelligence, reliability, flexibility, complexity, and cost. Each main criterion has several sub-criteria. The design process using MMP includes a fuzzy aggregation function based on Choquet fuzzy integrals which can efficiently model the interdependencies between a subset of criteria. Although, the main difficulty of the Choquet method is the identification of its fuzzy measures which exponentially increase by the number of design objectives. The objective of this study was to provide a framework to support the designers with the identification of fuzzy measures based on various available information and design preferences. We discussed three different methods of fuzzy measure identification applied to a case study of the conceptual design of a vision-guided quadrotor drone. These methods include using a Sugeno fuzzy model, a leaning data set, and fuzzy semantics. The results obtained from each method have been presented in the case study section and finally, a discussion on each method and their applications was carried out. From the implementation and results, we infer that in the case that there is not enough information about the design preferences or the interaction and importance of coalitions of criteria, using Sugeno $\lambda$-measures can be an abstract and efficient way. When only the relative global scores on each design alternative are available, the identification based on learning data is shown to be effective since this method does not need any information on importance and interaction indices. The data sets can be obtained from previous design cases or from an available database. This suggests an interesting subject of future work where the implementation of a web-based integrated platform connecting various design projects would be explored. In the absence of the global scores, the method combining the fuzzy measure semantics and learning data can be used. This method calls only for ordinal information on the alternatives and their importance of the criteria.